\RequirePackage{fix-cm}
\documentclass{svjour3}                     
\smartqed  

\RequirePackage[hyphens]{url}
\usepackage{appendix}
\usepackage{amsmath}
\usepackage{graphicx}
\usepackage{lineno}
\usepackage{array}
\usepackage{longtable}
\usepackage{natbib}

\usepackage[utf8]{inputenc}
\usepackage{tikz,amsmath}
\usetikzlibrary{positioning}
\PassOptionsToPackage{hyphens}{url}\usepackage{hyperref}
\usepackage{graphicx}
\usepackage{amsmath,blkarray}
\usepackage[ruled,vlined]{algorithm2e}
\usepackage{amssymb}

\usepackage{lineno}

\usepackage{amsmath,amssymb,enumitem}

\linenumbers
\newcommand*\patchAmsMathEnvironmentForLineno[1]{%
\expandafter\let\csname old#1\expandafter\endcsname\csname #1\endcsname
\expandafter\let\csname oldend#1\expandafter\endcsname\csname end#1\endcsname
\renewenvironment{#1}%
{\linenomath\csname old#1\endcsname}%
{\csname oldend#1\endcsname\endlinenomath}}%
\newcommand*\patchBothAmsMathEnvironmentsForLineno[1]{%
\patchAmsMathEnvironmentForLineno{#1}%
\patchAmsMathEnvironmentForLineno{#1*}}%
\AtBeginDocument{%
\patchBothAmsMathEnvironmentsForLineno{equation}%
\patchBothAmsMathEnvironmentsForLineno{align}%
\patchBothAmsMathEnvironmentsForLineno{flalign}%
\patchBothAmsMathEnvironmentsForLineno{alignat}%
\patchBothAmsMathEnvironmentsForLineno{gather}%
\patchBothAmsMathEnvironmentsForLineno{multline}%
}

\begin{document}

\nolinenumbers

\title{Solving reward-collecting problems with UAVs: a comparison of online optimization and $Q$-learning \thanks{R.Y. is partially supported by NSF DMS 1916037 and Consortium for Robotics and Unmanned Systems Education and Research (CRUSER).}
}

\titlerunning{Solving reward-collecting problems with UAVs}        

\author{Yixuan Liu         \and
        Chrysafis Vogiatzis \and
        Ruriko Yoshida \and
        Erich Morman 
}


\institute{R. Yoshida \at
              Operations Research Department\\ Naval Postgraduate School\\ 1411 Cunningham Road\\ Monterey, CA 93943-5219\\
              Tel.: +1-831-656-2973\\
              Fax: +1-831-656-2595\\
              \email{ryoshida@nps.edu}           
           \and
           Y. Liu \at
           Operations Research Department\\ Naval Postgraduate School\\ 1411 Cunningham Road\\ Monterey, CA 93943-5219\\
           \and
           C. Vogiatzis \at
              Industrial and Enterprise Systems Engineering\\ University of Illinois at Urbana-Champaign\\ Urbana, IL 61801-3080\\
              \and
            E. Morman \at
            Defense Resource Management Institute\\ Naval Postgraduate School\\ 699 Dyer Road, Monterey, CA 93943-5219\\
}

\date{Received: DD Month YEAR / Accepted: DD Month YEAR}

\maketitle

\begin{abstract}
Uncrewed autonomous vehicles (UAVs) have made significant contributions to reconnaissance and surveillance missions in past US military campaigns. As the prevalence of UAVs increases, there has also been improvements in counter-UAV technology that makes it difficult for them to successfully obtain valuable intelligence within an area of interest. Hence, it has become important that modern UAVs can accomplish their missions while maximizing their chances of survival. 
  In this work, we specifically study the problem of identifying a short path from a designated start to a goal, while collecting all rewards and avoiding adversaries that move randomly on the grid. \textcolor{black}{We also provide a possible application of the framework in a military setting, that of \textit{autonomous casualty evacuation}.} We present a comparison of three methods to solve this problem: namely we implement a Deep $Q$-Learning model, an $\varepsilon$-greedy tabular $Q$-Learning model, and an online optimization framework. Our computational experiments, designed using simple grid-world environments with random adversaries showcase how these approaches work and compare them in terms of performance, accuracy, and computational time.
\end{abstract}

\section{Introduction}

Due to the technological advances made in the past decades and numerous other factors, the US military has recognized the increased value of UAVs in modern warfare campaigns \citep{NAP11379}. UAVs are cost-effective, safe for operators, and can provide surveillance and reconnaissance in areas that might otherwise be inaccessible. However, with the increased usage and threat of UAVs, counter-UAV technology has become an increasing concern when considering UAV deployment. The US military, for example, is investing hundreds of millions of dollars to develop jamming, laser, and gun technologies that will serve to defend against UAV threats. \citep{hoehn_sayler_2020} In this paper, we explore three potential methods that can be used for UAV path planning for surveillance missions when there is an active adversary trying to capture or destroy our UAVs. We develop a simple grid environment game that is composed of rewards that our UAV agent must capture and adversaries whom our UAV agent must avoid while traversing through the environment in the shortest path possible. We will attempt the game with the three methods and compare the results. 

According to \citet{survey}, robotic systems use five main deliberation functions in order to fulfill their missions. The five deliberation functions are planning, acting, monitoring, observing, and learning. \textcolor{black}{In this work, we focus on exploring the planning function of robotic systems; specifically we pose it as a probabilistic planning problem.} We propose here to investigate the use of online optimization, $\varepsilon$-greedy tabular $Q$-Learning, and reinforcement learning as tools to gain insight into better UAV movement patterns by solving a dynamic, stochastic, rewarding-collecting path problem with side constraints. The UAV is sent from a source node $s$ to a terminal node $t$ in a network; its goal is to reach the terminal node in a fast and safe manner. \textcolor{black}{We define fast as using a short number of hops or links in the network; we define safe as avoiding a series of adversaries that are deployed on that same network.} In addition, whilst routing, the UAV collects useful information by visiting certain nodes of the network. This is the reward-collecting part of the problem. The dynamic and stochastic nature of the problem comes from the presence of adversaries that may move in a deterministic or stochastic fashion through the nodes in an effort to capture our agent. 

Our online optimization method is a greedy algorithm to estimate the best path over the entire time horizon by sequentially solving simple linear programming problems at each time step. Because adversaries are moving with unknown (stochastic or deterministic) patterns, the problem over the entire time horizon is stochastic.  However, in our method, we set up a deterministic problem at each time step using the estimated transition probabilities of adversaries. Then, we use linear programming to find an optimal step for the UAV to move during the next time step by minimizing the risk to encounter adversaries and to take an extra step towards the rewards (and the exit) at the same time.  This is a sequential optimization problem since the optimal solution at the current time step sets up the linear programming problem in the next time step. This happens using the most up-to-date transition probabilities of adversaries. Finally, it is a greedy algorithm since we take the optimal solution at the current time step (local optimal solution), but not the global optimal solution in the entire time horizon.

\citet{faust} conducted similar work on using reinforcement learning to learn an optimal policy for cargo-bearing UAVs to deliver goods. Their work consisted of breaking down a large action space and complex environment into smaller and simpler problems for efficient learning. We take the opposite approach in this paper by starting with an extremely simple problem and eventually scaling up. We will begin experimentation on a $5 \times 5$ grid environment with a single agent, reward, and adversary. 
Then we compare to an $\epsilon$-greedy tabular $Q$-Learning model implemented in {\tt Python} as well as to a Deep $Q$-Learning model implemented using the {\tt Python} package ``keras'' \citep{chollet2015keras} and ``tensorFlow'' \citep{tensorflow}.  

{\color{black}
\subsection{Contributions}
In this paper, we propose an online optimization model on a shortest path and reward collecting problem with a random adversary.  Especially, we compared $\epsilon$-greedy tabular $Q$-learning method, deep $Q$-learning method, and online optimization model on a specific dynamic and stochastic game with full information: the agent takes a shortest path with collecting all rewards while it avoids the adversary which moves certain patterns, namely clock-wise, counter clock-wise, or random moves. In this game, the agent and adversary can access to full undated information.  Simulated experiments show that $\epsilon$-greedy tabular $Q$-learning method and deep $Q$-learning method work very well on games with clock-wise and counter clock-wise adversaries, while they have difficulties to find all rewards on a game with a random adversary.  On the other hand, our online optimization model works very well on games with clock-wise, counter clock-wise, and random adversaries.    Finally, we propose an application of this model to {\em Autonomous Casualty Evacuation (ACE)} proposed by Dr.~Timothy Bentley at Office of Naval Research (ONR).
}


{\color{black}
\subsection{Autonomous Casualty Evacuation}

\sloppy

Before proceeding to describe our work, we provide a possible application of the setup to a real-life military UAV problem. More specifically, we focus on the problem of \textit{autonomous casualty evacuation} or \textit{extraction} (see, e.g., \cite{ACE1} or \cite{ACE2}). For an excellent overview of the problem and its many facets, we refer the interested reader to \cite{williams2019review}. The problem has been of interest to the Office of Naval Research (ONR) and the Naval Force Health Protection program (\url{https://www.onr.navy.mil/en/Science-Technology/Departments/Code-34/All-Programs/warfighter-protection-applications-342/force-health-protection}).

In our case, we consider a setup as in Figure \ref{ACE_figure}. In this setup, we assume the existence of casualties in need of evacuation from an adversarial territory. The adversary is aware of the location of the casualties; however, the adversary only patrols the area in anticipation of reinforcements. At the same time, our agents have the simple goal of reaching the casualties and removing them from adversarial territory fast and safely. We finally make one assumption about the behavior of the adversary: absent any information about their strategy, we can assume that their location at any point is stochastic and is guided by a (unknown to us) probability distribution. 

\begin{figure}
{\color{black}
\centering
\begin{tikzpicture}[scale=1,every node/.style={minimum size=1cm},on grid]
    \begin{scope}[
            yshift=-83,every node/.append style={
            yslant=0.5,xslant=-1},yslant=0.5,xslant=-1
            ]
        \fill[white,fill opacity=0.9] (0,0) rectangle (5,5);
        \draw[step=4mm, black] (0,0) grid (5,5); 
        \draw[step=1mm, green!50,thin] (3,3) grid (2,2); 
        \draw[step=1mm, red!50,thin] (1.5,3.4) grid (2,1.5); 
        \draw[step=1mm, red!50,thin] (1.5,3) grid (3.5,3.4); 
        \draw[step=1mm, red!50,thin] (3,3) grid (3.5,1.5); 
        \draw[step=1mm, red!50,thin] (3.5,1.5) grid (1.5,2); 
        \draw[red!50,very thick] (1.5,3.4) rectangle (3.5,1.5); 
        \draw[black,very thick] (0,0) rectangle (5,5); 
        \fill[blue] (0.05,0.05) rectangle (0.35,0.35); 

    \end{scope}
    	
    \draw[-latex,thick,green!50!black](2.7,-1.9)node[right]{Casualty location}
        to[out=180,in=90] (0,-.5);
        
     \draw[-latex,thick,red!50!black](2.3,1.9)node[right]{Adversary patrol territory}
        to[out=180,in=90] (0,.35);
        
    \draw[-latex,thick,blue](-2,-3)node[left]{Initial UAV location}
        to[out=0,in=230] (-0.2,-2.9);

\end{tikzpicture}
\caption{A pictorial example of the autonomous casualty evacuation problem that motivates our work. On the figure, we present (i) the adversary patrol territory, in which the adversary can be; (ii) the casualty location which we need to reach safely and fast; (iii) the starting position of our UAV. The goal is to reach a point in the periphery after visiting the casualty location without being intercepted from the patrolling adversary. \label{ACE_figure}}}
\end{figure}
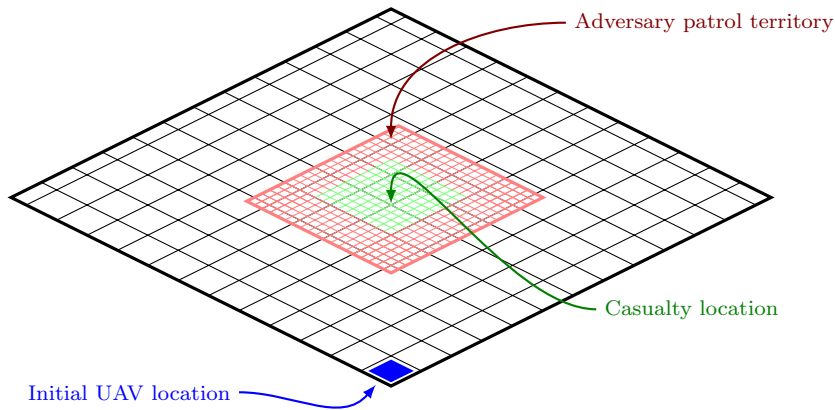

We can view this last assumption as the existence of an ``adversary cloud''. That is, at any given point and based on the previously known adversarial position, the adversary patrolling around the casualty location can be at any other position following a probability distribution function. In our experimental setup, we will only consider adversaries that can move in the vicinity of their previously known positions. That said, our work can be generalized to other probability distributions.}

\section{Background}

In this section we present some of the necessary background and fundamental information behind Deep $Q$-Learning, $\varepsilon$-greedy Tabular $Q$-Learning, and online optimization. 

\subsection{Machine Learning}

To understand how $Q$-Learning works, we must first discuss the machine learning umbrella that $Q$-Learning falls under. Machine learning is a subfield of computer science that describes the process of solving problems by gathering a relevant dataset and algorithmically using the dataset to build a statistical model \citep{machinelearning}. Machine learning is generally categorized into three types: supervised learning, unsupervised learning, and reinforcement learning. Supervised learning consists of training a model with a dataset full of labeled examples to be able to categorize inputs into known classes. Unsupervised learning consists of training a model with a dataset full of unlabeled examples and categorizing inputs based off of features from the unlabeled dataset. Reinforcement learning, which $Q$-Learning is an example of, consists of learning a policy on how an agent should act based on the state of its environment in order to maximize some kind of reward system. 

\subsection{Reinforcement Learning}

We focus on solving our specific problem with a reinforcement learning algorithm in this paper because of reinforcement learning's ability to solve near-optimally, complex and large-scale Markov decision processes on which classical dynamic programming breaks down \citep{Gosavi}. While our current problem is small in size and can be solved with ease through other methods, we investigate solving it through reinforcement learning as real life problems involving UAVs in unfamiliar environments will be much greater in complexity.

Reinforcement learning is an area of machine learning that focuses on how an agent should behave inside of an environment. Different actions taken by the agent in different states might have positive, neutral, or negative outcomes in the form of rewards. The goal of reinforcement learning is to derive a policy of actions the agent should take to maximize overall reward. 

The process of finding an optimal policy through reinforcement learning requires four elements besides an agent and environment:``a policy, a reward signal, a value function, and, optionally, a model of the environment" \citep{Sutton}. The \textbf{policy} $\pi$ can be a lookup table or a function that allows the agent to know which action will produce the highest reward based on the state of the agent. The \textbf{reward signal} tells us whether an action is a "good" action or a "bad" action based on the rewards we receive. The overall goal of reinforcement learning is to maximize the long term rewards earned by our agent through its actions. The \textbf{value function} differs from the reward signal in that while the reward signal is the immediate reward gained from taking actions, the value of a state refer to the long term desirability from being in that state and the subsequent states that follow. The \textbf{model} element is not a necessity, as we can conduct reinforcement learning whether or not we have a model of the environment. A model of the environment should allow us to make predictions about future states and rewards based on current states and actions. When we do not have a model of the environment, reinforcement learning takes a completely trial and error approach to learn more about the environment.

\subsection{$Q$-Learning}

For our problem, we will use a reinforcement learning algorithm called $Q$-Learning \citep{Watkins}, where the $Q$ is understood to mean ``quality". $Q$-Learning proceeds as following:``an agent tries an action at a particular state, and evaluates its consequences in terms of the immediate reward or penalty it receives and its estimate of the value of the state to which it is taken. By trying all actions in all states repeatedly, it learns which are best overall, judged by long-term discounted reward" \citep{Watkins}.

Let $s$ represent the state of an environment the agent is in and $a$ be the action the agent takes from state $s$, the goal of $Q$-Learning is to be able to calculate the $Q$ value, $Q(s,a)$, for all state action pairs in the environment. An optimal policy can then be found by always picking the action with the highest $Q$ value in every state. Given a policy $\pi$, $Q(s,a)$ can be thought of as the the immediate reward gained from taking action $a$ in state $s$ plus the discounted reward of following policy $\pi$ in the future. Calculating the $Q$ values for an environment can be quite complex as it would require taking every possible action in all possible states for a finite Markov decision process. In $Q$-Learning, we will use Bellman's equation to approximate the $Q$ values for an optimal policy:
\begin{equation}
Q(s,a) = r(s,a) + \gamma \underset{a'} {\mathrm{max}} Q(s',a'),\\
\label{Bellman}
\end{equation}
where $r(s,a)$ is the immediate reward of taking action $a$ in state $s$, $Q(s',a')$ is the $Q$ value function for the next state $s'$ and next action $a'$, and \(\gamma\) is the discount factor to used to ensure future rewards are worth less than the immediate reward. From Bellman's equation, we calculate $Q(s,a)$ by adding the immediate reward from taking taking action $a$ in state $s$ and the $Q$ value of the next state action pair with the highest $Q$ value. If we create a policy based on always choosing state action pairs with the highest $Q$ values generated with Bellman's equation, it can be proven that the policy will eventually converge to the optimal policy \citep{Watkins}. Once we have our $Q$ values, we will have our agent choose to either make the move that results in the highest $Q$ value or make a move completely at random. This exploitation versus exploration strategy allows us to constantly seek new paths that might result in higher $Q$ values. 
 
 \subsection{$\varepsilon$-greedy Tabular $Q$-Learning}
Our initial $Q$-Learning model leverages a variant of Bellman's equation to generate feasible decision policies $\pi$ for our grid environment. The rationale for the tabular designation of this approach is that the iterative $Q$-Learning process generates a look-up table of $Q$ values associated with each state-action pairing $Q(s,a)$ that exists in our decision environment.  At each iteration $n$ in the training process of our tabular model, we take an action $a$ according to

\begin{equation}
a^n =  \underset{a\epsilon A} {\mathrm{arg max}} \bar{Q}^{n-1}(s,a).\\
\label{action selection}
\end{equation} 
where $\bar{Q}^{n-1}(s,a)$ is the $Q$ value as of the $n-1^{th}$ iteration.

Two good resources that elaborate on the $\epsilon$-greedy tabular $Q$-Learning method in more detail include the work by \citet{Powell} and by \citet{Sutton}. Unfortunately, one of the challenges of this method is that it suffers from the curse of dimensionality.  An effective training process of the tabular $Q$-Learning approach will require the model to sample each state-action pairing a sufficient number of times in order to learn its $Q$ value.  For each additional state or action that is included in our decision system, there is exponential growth in the computational complexity of the $Q$-Learning process making it difficult to apply this approach to even the simplest of problems.  Therefore, we need to investigate alternative solution approaches such as Deep $Q$-Learning and online optimization.  However, for manageable problems such as our adversary Grid environments, the tabular approach has value in serving as an initial baseline method to the more advanced methodologies.

 \subsection{Deep $Q$-Learning}
Our Deep $Q$-Learning model utilizes a neural network with more than 1 hidden layer, thus this $Q$-Learning algorithm is termed Deep $Q$-Learning. This technique is motivated by a company named Deepmind that has gained a lot of attention in recent years due to the success of their Deep reinforcement learning methods. Deepmind proposed a method called Deep $Q$ Network (DQN) to perform a variant of $Q$-Learning with convolutional Deep neural networks to keep track of future rewards (see \citet{Mnih}).  DQN uses a neural network function approximator called a $Q$-network that can be trained by minimising a sequence of loss functions by stochastic gradient descent. The motivation behind this method is that traditional methods of iteratively updating the $Q$ values for an environment with Bellman's equation has very high computational complexity. DQN is able to more efficiently generate $Q$ values by training the neural network with random batches of experience replays that are composed of $e_t = (s_t, a_t, r_t, s_{t+1})$. These experience replays are stored every time the agent takes an action and a random sample of them are used to train the neural network every time step. Their method has achieved exceptional results when playing Atari games and their success led to their acquisition by Google for \$500 million \citep{Shu}. DQN has been shown to work well in a diverse array of applications, including but not limited to stock market forecasting \citep{carta2020multi}, energy-efficient scheduling schemes \citep{zhang2017energy}, and handwritten digits recognition \citep{qiao2018adaptive}. 

\subsection{Online Optimization}

The final approach we will apply to our problem is online optimization. While traditional statistical learning methods are robust, they are often unable to solve problems problems where the data is dynamic and our knowledge of the problem is incomplete. Online optimization provides us with a way to solve sequential problems where we make no probabilistic assumptions of the distributions of our inputs and require only partial knowledge or a partial view of the problem \citep{bubeck_2011}. Online optimization can be used to solve a variety of problems such as network routing \citep{he2013endhost}, e-mail spam filtering \citep{wang2006svm}, or stock portfolio selection \citep{li2014online}. The general idea behind online optimization is that we are solving a series of problems in a sequential manner, instead of one big problem. \textcolor{black}{Each of the problems is solved \textit{in sequence} at each time step based on information we obtained from solving earlier problems (i.e., from previous time steps) and through observation of the resultant environments.}

A generic online optimization decision process for an optimizing agent can be described as the following \citep{online_formulation}. Given a time horizon $T=\left\{0,1,\ldots, |T|\right\}$, for every time step $t\in T$:

\begin{enumerate}[itemindent=1cm,label=\bfseries Step \arabic*.]
  \item The agent chooses an action $x_t$ based on its current knowledge of the problem after solving an optimization problem $LP(t)$.
  \item The agent incurs some sort of loss based on the selected action, $l_t(x_t)$. 
  \item The agent observes the new environment and obtains new knowledge. 
\end{enumerate}

In the process described above, the agent's action $x_t$ must always come from a finite amount of possible actions and the loss function $l_t$ can be of any form. The goal of online optimization is ultimately to minimize the loss function as we proceed through the time steps. 




\section{Experimental Setup}

In this section, we describe the experimental setup and the specific implementation details of the approaches adopted.

\subsection{Game Setup}

A simple set up of the game in a $5\times 5$ grid environment with a single reward and adversary can be seen in Figure \ref{fig:5x5setup}:

\begin{figure}[htp]
    \centering
    \includegraphics[width=4cm]{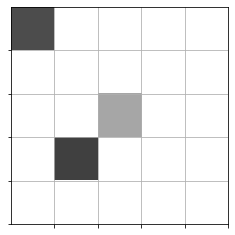}
    \caption{$5\times 5$ Example Initial Setup}
    \label{fig:5x5setup}
\end{figure}

In Figure~\ref{fig:5x5setup}, we have our agent starting in the upper left corner, our reward in the center of the grid, and our adversary in the cell to the lower left of our reward. The adversary can only move in the cells directly surrounding our reward, but our agent can move anywhere in the grid. The goal of the agent is to traverse through the grid and reach the reward in the center of the grid using the shortest path while simultaneously avoiding the adversary. Once it has reached the reward, it then proceeds to the lower right corner in order to finish the game in a similar fashion of finding the shortest path and simultaneously avoiding the moving adversary.

\subsection{Adversary Types and Actions}
The environment for the adversary is fixed, and consists of a subset of the entire environment.  \textcolor{black}{This means that the adversary can only visit a subset of the entire grid (see the patrolling area in red in Figure \ref{ACE_figure}) and these locations are fixed throughout each run.} All rewards in our environment are surrounded by a moving adversary. \textcolor{black}{All adversary types considered can only move to cells that are adjacent to their current position.} That said, different adversaries can move in differently: (i) clockwise around the reward; (ii) counterclockwise around the reward; or (iii) randomly around the reward. In each time step of our game, our agent will take one step, and then our adversary will take one step. At the end of each time step, if our agent and adversary end up in the same node, then we consider the agent to be captured and the game to be over. Finally, every adversary can take only four moves possible depending on their location in the environment: that is, they can move (i) up; (ii) down; (iii) left; or (iv) right in the given subset of the entire grid. 

\subsection{$\epsilon$-greedy Tabular $Q$-Learning Method}

The tabular $Q$-Learning algorithmic process is described in \citep{Sutton} and uses the following update process for the $Q$ values after each one-step iteration of the model:          
\begin{equation}
Q(s,a) \leftarrow{} Q(s,a) + \alpha[r(s,a) + \gamma \underset{a'} {\mathrm{max}} Q(s',a')-Q(s,a)].\\
\label{Q_Value_Update}
\end{equation} 
The update structure is similar to Bellman's equation \eqref{Bellman}.  However, here we are updating the $Q$ values with a weighted value of the reward for taking action $a$ and the difference between the discounted maximum $Q$ value of the next state $s'$ and the current state $Q$ value.  The $\alpha$ term is an assigned learning rate.   

Once the training process is completed, we create a decision policy from our look-up table by selecting the action $a$ that gives the maximum $Q$ value for a given state $s$:
\begin{equation}
a =  \underset{a\epsilon A} {\arg\max} \bar{Q}(s,a).
\end{equation}

As described in \citep{Powell}, the problem with this formulation is that it will only train state-action pairs with the highest $Q$ values and overlook other potentially good state-action pairings.  Therefore, as our algorithm moves through the training process, it needs to create a balance between exploiting state-action pairings with the highest $Q$ values and exploring other potentially good state-action pairings.  This is done by creating an $\epsilon$-greedy rule where with a probability of 1-$\epsilon$ we choose an action according to \eqref{action selection} and with a probability of $\epsilon$ we randomly pick an action. 

In order to balance between exploration and exploitation during the training process we systematically decline the $\epsilon$ after each epoch, starting with an $\epsilon$ value near one and ending with a value near zero.  Thus, initial epochs during the algorithmic training process will favor exploration steps while the latter epochs favor exploitation steps.  We will use an adapted deterministic harmonic epsilon-decay rate as suggested by both \cite{Darken} and \cite{Gosavi_2}:
\begin{equation}
    \epsilon_n = \frac{\epsilon_{n-1}}{1+\frac{n^2}{\beta + n}} \\
\label{epsilon decay}
\end{equation}

In \eqref{epsilon decay}, $n$ is the epoch number and $\beta$ is a tuning parameter.   

\subsection{Deep $Q$-Learning Method}

Our Deep $Q$-Learning model and code was built directly upon the Tour De Flags code \citep{zafrany} written by Samy Zafrany. Most of the parameterization of our reinforcement learning and neural network model remain the same as his as those have already been optimized for a similar experiment. 

As we have described earlier in Section 2, our Deep $Q$-Learning algorithm relies on finding the the optimal policy by always choosing state action pairs with the highest $Q$ values generated with Bellman's equation. The trick is being able to generate correct the correct $Q$ values from our agent's experiences exploring the game environment. The method we used to do this is to create a model where the target function of our neural network to match the right side of Bellman's equation. Our goal is to minimize the difference between our neural network's estimation of the state action $Q$-values generated by the right side of Bellman's equation, and the actual $Q$-values our agent experiences from traversing through our game environment. 

We train our neural network by allowing our agent to traverse through the environment and keeping a record of the its most recent experiences in a way that we will explain more below. Each time it moves, it will also select a random sample of the record of its most recent experiences and use this information to train our neural network. Once the agent's record of most recent experiences has grown too large, we will delete the older records and only keep the newer ones. This allows our neural network to be trained with newer data as our agent learns more and more of the environment. 

Some of the parameter values that we use in our Deep $Q$-Learning algorithm are listed below:
\begin{itemize}
    \item number of epochs - number of times our agent attempts the game: 1000
    \item max memory - the number of records of recent experiences our agent keeps before deleting older records: 500
    \item data size - the number of records of recent experiences used to train our neural network each time the agent moves: 100
    \item exploration factor - the percentage of times the agent chooses a completely random move instead of an optimal move: 0.2
    \item discount factor - the $\gamma$ coefficient in the right side of Bellman's equation used to discount the worth of future rewards - 0.97
\end{itemize}

We will conduct reinforcement learning in this paper in {\tt Python} using Tensorflow and Keras packages. Originally developed by Google, TensorFlow is an open-source platform to facilitate the development and application of machine learning techniques \citep{tensorflow}. Similarly, Keras is an API to enable deep learning approaches on top of TensorFlow with the goal of fast prototyping \citep{chollet2015keras}.

Keras works by processing vectorized and standardized representations of raw data such as NumPy arrays. Each time the agent takes a step, often referred to as an episode, we store data in the following manner: episode = [env\_state, action, reward, next\_env\_state, game\_over]. Each element of the array are either numbers or arrays of numbers representing respectively the game environment before any action takes place, the action the agent takes, the reward that resulted from the respective action, the updated game environment after the action takes place, and whether or not the game is still ongoing. Since Keras only works with vectorized representations of raw data, the environment states recorded are compressed 1-dimensional arrays where each element represents a cell in our grid environment. Different values in the arrays allow the neural network to know where the agent and the reward are located. 

We use the same neural network that Zafrany uses in his Tour de Flags code, which is a sequential Deep neural network model built with Keras. The neural network model has four fully connected layers composed of one input layer, two hidden layers, and 1 output layer. Our input layer is the same size as our grid environment, so a $5\times 5$ grid environment will have an input layer of 25. This is because our input to the neural network is a one dimensional array that displays where within the grid environment our agent is currently located. Our output layer is size 4, representing the 4 actions our our agent can take and producing a $Q$-value for each action.

We use LeakyReLu activation layers in our neural network to prevent our neural network from becoming too sparse. The LeakyReLu class in keras is the leaky version of a rectified linear unit, which allows a small gradient when the unit is not active instead of the usual gradient of zero \citep{leakyrelu}:

\begin{equation}
  f(x)=\begin{cases}
    \alpha \cdot x, & \text{if $x<0$}.\\
    x, & \text{if $x >= 0$}.
  \end{cases}
\end{equation}The optimizer we use in our neural network is the Adam optimizer \citep{adam}, which uses a stochastic gradient descent method.

\subsection{Online Optimization Method}
Our online optimization problem is set up where our agent is able to observe the adversary for a certain number of time steps in order to obtain an observed transition probability matrix for the adversary's future moves. At each time step, the agent knows the current location of the adversary and solves a linear programming problem to calculate the optimal path to win the game based on the observed adversary transition probability matrix and current adversary location. Our online optimization model is described below.

Let $G(V,E)$ be an undirected graph/grid with a series of targets on the grid $k\in\mathcal{K}$, where $\mathcal{K}\subseteq V$. Let $r_{k}^t$ be the reward from reaching target $k$ at time $t\geq t_0$, where $t_0$ marks the current time; for convenience, we let $r_{i}^{t}=0, \forall i\in V\setminus \mathcal{K}$. We consider one of the locations in the grid the ``exit'' node, $d$. When the agent reaches this node, they have effectively exited the grid, and they no longer need to move.

Moreover, define $p_{i}^t$ as the probability of seeing an adversary at location $i$ at time $t\geq t_0$. Finally, let $\phi$ be the ``importance'' we place in avoiding the adversary.

We are now ready to move to the definition of our decision variables. We define two integer variables, $x_{ij}^t$ and $y_{ik}^t$, as follows:

\begin{align*}
    & x_{ij}^t=\begin{cases} 1, & \text{if traversing $\left(i,j\right)\in E$ at time $t\geq t_0$,} \\ 0, & \text{otherwise.} \end{cases}\\
    & y_{i}^t=\begin{cases} 1, & \text{if arriving at node $i\in V$ at time $t\geq t_0$,} \\ 0, & \text{otherwise.} \end{cases}\\
\end{align*}

Before we describe the mathematical formulation, we describe the process. At each time step $t_0$ and starting from $t_0=0$, we solve this optimization problem. Then, we observe the movement of the adversary and then move again. In effect, we collect a series of optimal solutions, one for each current time $t_0$. We then have the formulation in \eqref{formulation}. For each time step $t_0=0, 1, \ldots$, we have:

\begin{subequations}
    \label{formulation}
    \begin{align}
    \label{objective}
         \min~~ & \sum\limits_{i\in V}\sum\limits_{t\geq t_0} \left(t-r_{i}^t+\phi\cdot p_{i}^t\right)\cdot y_{i}^t \\
         \label{assignment1}
        s.t.~~ & \sum\limits_{i\in V\setminus \left\{d\right\}} y_{i}^{t}\leq 1, & \forall t\geq t_0, \\
        \label{assignment2}
        & \sum\limits_{t\geq t_0} y_{i}^{t}\leq 1, & \forall i\in V\setminus \left\{d\right\}, \\
         \label{preservation1}
        & \sum\limits_{j:(i,j)\in E} x_{ij}^{t+1} =y_{i}^{t}, & \forall i\in V\setminus\left\{d\right\}, \forall t\geq t_0, \\
         \label{preservation2}
        & \sum\limits_{j:(j,i)\in E} x_{ji}^{t} =y_{i}^{t+1}, & \forall i\in V, \forall t\geq t_0, \\
        \label{exit_constraint}
        & \sum\limits_{t\geq t_0}\sum\limits_{i:\left(i,d\right)\in E}x_{id}^t=1, \\
        \label{binary1}
        & x_{ij}^t\in\left\{0,1\right\}, & \forall \left(i,j\right)\in E, \forall t\geq t_0,\\
        \label{binary2}
        & y_{i}^{t}\in\left\{0,1\right\}, & \forall i\in V, \forall t\geq t_0.
    \end{align}
\end{subequations}

In the formulation, the objective function in \eqref{objective} minimizes a ``risk function'' for our agent. The risk function consists of three components, all multiplied by the decision variable of visiting some node $i$ at some future time $t$: \begin{enumerate}
    \item a ``negative'' reward, as we would like to collect as many rewards as possible;
    \item a $\phi\cdot p_i^t$ component, which aims to consider future positions the agent may take; and
    \item a multiplication by $t$, since we would like the agent to take fewer steps on their way to collecting the rewards and exiting the grid.
\end{enumerate} 

Constraints \eqref{assignment1} and \eqref{assignment2}, we simply enforce that our agent is in at most one location at each time; observe how this allows for being in ``no location'' if the agent has reached the exit point $d$. Constraints \eqref{preservation1} and \eqref{preservation2} are flow preservation constraints which state that the agent has to use one of the edges around them at each location they find themselves. Finally, constraint \eqref{exit_constraint} enforces that the agent will have to exit at some point; recall that the objective also makes the agent want to exit faster, if possible. Constraints \eqref{binary1} and \eqref{binary2} are variable restrictions seeing as both of our decision variables are binary. Our online optimization code is implemented and solved in {\tt Python} using the Gurobi Optimizer \citep{gurobi}.

\section{Computational Results}

\begin{figure}[!htp]
    \centering
    \includegraphics[width=4cm]{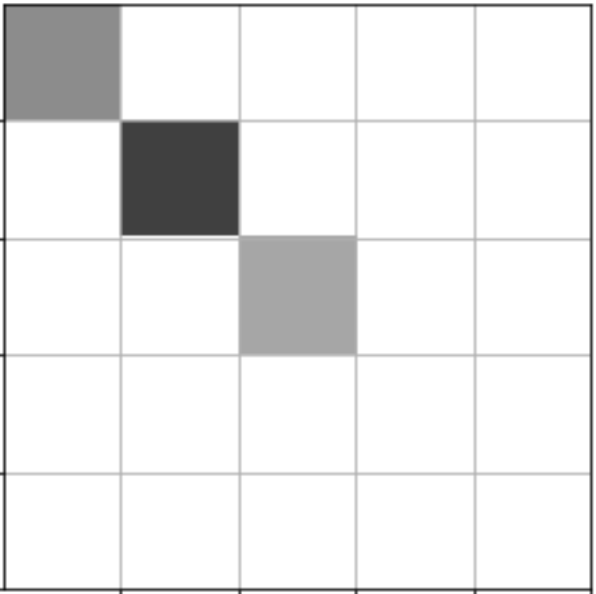}
    \caption{$5\times5$ Example Initial Setup}
    \label{fig:example5x5setup}
\end{figure}

We begin the section with a description of the initial $5\times 5$ setup. In Figure~\ref{fig:example5x5setup}, we can see the initial placement of our agent, adversary, and reward. The agent starts in the upper left corner of the grid, the reward is in the center of the grid, and the adversary is to the upper left corner northwest of the reward. The agent must find a short and safe path to capture the reward and reach the lower right corner of the grid. 

We begin our experiments with our reinforcement learning algorithm. We set up our experiments for three different types of adversary that move clockwise, counterclockwise, and randomly. We initialize our game with a $5\times 5$ grid environment, a single reward in the center of the $5\times 5$ grid, and a single adversary that starts in the upper left cell adjacent to our reward. The rewards for our game are set up as below.
\begin{itemize}
    \item Each step that does not result in capturing a reward or reaching the final destination: $-1$
    \item Reaching the reward: $+200$
    \item Being captured by the adversary: $-1000$
    \item Reaching the bottom right corner of our $5\times 5$ grid having captured the reward: $+100$
\end{itemize}

Since there is only one adversary at most surrounding the reward, there is no way for the adversary to block off the agent in a way that the agent is unable to complete the game in 8 steps, which is the minimum number of steps needed to reach both the reward and the bottom right corner from the upper left corner. Using the reward system described above, taking 8 steps to complete the game will result in a final reward of 294, which we will consider to be optimal for our $5\times 5$ experiments. 

Initial experimentation of the $5\times 5$ grid model involved using the $\epsilon$-greedy tabular $Q$-Learning method.  For each adversary movement type, we ran the model for 1,000 epochs and compared the results of ten different trials. Table \ref{table:1} summarizes the findings.

\renewcommand*{\arraystretch}{1.3}
\begin{table}[!ht]
\centering
\begin{tabular}{ |p{5.5cm}||p{1.5cm}|p{1.5cm}|p{1.5cm}|  }
 \hline
 \multicolumn{4}{|c|}{$5\times 5$ Environment - $\epsilon$-Greedy Tabular Method} \\
 \hline
Adversary Movements  & Clockwise & Counter- & Random\\
 & & clockwise & \\
 \hline
\# of Successes   &  7   & 6 & 0\\
Avg. \# of Wins if Successful & 387.7 & 434.4 & N/A\\ 
Optimal Successes & 5 & 6 & 0\\
Avg. \# of Optimal Wins if Optimal & 313.2 & 250.5 & N/A\\  \hline
Avg. \# of Wins If Not Successful & 264.0	& 170.8	&	122.0\\
Avg. \# of Optimal Wins If Not Successful &  74.0 & 31.8 & 14.0  \\
\hline
Time(secs.) If Successful & 0.97	&	1.02	&	N/A\\
Time(secs.) If Not Successful & 16.58 & 25.20 & 16.07 \\
 \hline
 \end{tabular}
 \caption{$5\times 5$ Environment: $\epsilon$-Greedy Tabular $Q$-Learning Results}
 \label{table:1}
\end{table}

\renewcommand*{\arraystretch}{1}

From Table \ref{table:1}, we see that this learning process had some moderate success with training our UAV against the clockwise and counterclockwise moving adversaries.  However, the algorithm struggled to discover a successful game winning route when facing an adversary that moves randomly.
While facing the clockwise moving adversary, the UAV found a successful route for seven of the ten trials, five of which were an optimal game completing eight-step path.  Furthermore, of the five non-optimal trials, four were only one $Q$ value away from having an optimal policy. When facing the counterclockwise adversary, the UAV was able to successfully find a successful path for six of the ten trials, where all six were also optimal eight-step pathways.  Also, of the remaining four non-optimal trials, all of these were only one $Q$ value away from having an optimal policy.  In summary, it appears that while facing an adversary that moves in a predictable manner, half of the time the $\epsilon$-greedy tabular $Q$-learning method is able to learn and generate an optimal look up table policy for moving our UAV through this environment and for the other half of the time, the look-up table policy is only one $Q$ value away from finding an optimal path.

Two interesting data points to compare from our findings are the \textit{average number of optimal wins if optimal} and \textit{the average number of optimal wins if not successful}.  It appears that on average, if the training process involved traversing the optimal path at least 250 times given the 1,000 epochs then there is a good chance our algorithm will produce an optimal look-up table.  If this number drops to as low as 75 or fewer, it is likely that the training process will not only struggle to find an optimal policy, it will likely not be able to even discover a feasible policy. 

Unfortunately, the $\epsilon$-greedy Tabular $Q$-Learning method did not perform well when facing the randomly moving adversary.  In this case, the learning process was not able to learn and create either an optimal pathway or a successful feasible pathway through the grid environment. The result is not surprising, given that for each of the 1,000 epoch training trials, an optimal pathway was traversed on average for only 14 times and a non-optimal feasible pathway traversed on average only 122 time. Nonetheless, one positive statistic that emerged from this scenario was that six of the ten generated tabular policies were only one Q value away from having an optimal pathway. 

Lastly, it was noted that when the algorithm was successful at generating a game competing feasible tabular policy, it would complete the 1,000 trials in nearly a second.  However, when the learning process was unable to generate a safe feasible tabular policy for our UAV, the computation time increased by factors of roughly 16 and 25.  In the latter cases, it appeared that the learning process would get stuck with cycling between adjacent squares generating large step counts to avoid the huge penalty of getting captured by the adversary, unaware of the positive trophy and exit points that could be obtained by moving through the pathway of the adversary.

Although there was mixed results from the performance of the $\epsilon$-greedy tabular $Q$-Learning algorithm, the approach does create baseline to now compare the two other methods examined in this research, the Deep $Q$-Learning algorithm and the online optimization algorithm.  

In an effort to make our Deep $Q$-Learning and online optimization algorithms more comparable, we do not have a training period for our Deep $Q$-Learning algorithm to train our neural network and then experiment with the trained neural network. We instead have each replication using our Deep $Q$-Learning algorithm consist of our agent attempting the game for 1000 epochs. If the agent successfully completes the game once, we consider the replication a success and end the replication. If the agent cannot successfully finish the game within 1000 epochs, then we consider the replication to be a failure. We set up our experimentation this way because each of our online optimization replications only allow our agent to attempt the game once. If we allow our reinforcement learning algorithm to have a training period, then we are allowing the agent to potentially win the game multiple times in the training period and thus unfairly comparing the two methods. While it is impossible to make comparisons between the two methods completely fair as they are intrinsically different algorithms and require different input, we attempt to make the two algorithms more comparable by stopping each Deep $Q$-Learning replication once the agent successfully completes the game once. 

Successfully completing a game requires the agent captures the reward, avoids the adversary, and reaches the lower right corner of our environment. If the agent is captured by the adversary at any point, then we consider the replication to be a failure. Each type of adversarial movement scenario is replicated 50 times. The average reward and regret are calculated only from replications that are successful in that the agent is able to capture the reward and reach the lower right corner of the grid without being captured by the adversary. 

\begin{table}[!ht]
\centering
\begin{tabular}{ |p{3cm}||p{2.3cm}|p{2.4cm}|p{2.3cm}|  }
 \hline
 \multicolumn{4}{|c|}{$5\times 5$ Environment - Deep $Q$-Learning Results} \\
 \hline
Adversary Movement  & Clockwise & Counterclockwise & Random\\
 \hline
\# of Successes   &  50   & 50 & 50\\
Average Reward & 292.6	& 292.56	&	292.6\\
Average Regret & 1.40	&	1.44&	1.40\\
Time(secs) & 173.74	&	198.60	&	312.74\\
 \hline
\end{tabular}
 \caption{$5\times 5$ Environment: Deep-$Q$ Learning Results}
 \label{table:2}
\end{table}

From Table \ref{table:2}, we can see that Deep $Q$-Learning performs well for the $5\times 5$ grid environment with a single adversary regardless of adversarial movement. The algorithm is able to successfully complete all 50 replications for each type of adversary movement. One thing to note is that it does seem to take longer for the algorithm to complete 50 replications when the adversary moves randomly. 

Now we will attempt to beat the game using online optimization. For our online optimization method, the game is simply a traditional optimization problem when the adversary moves clockwise or counterclockwise. As long as our agent can observe our adversary for 8 time steps before staring to move, it is able to know exactly how the adversary will move as it takes 8 steps for the adversary to complete its route and return to its starting position. The online optimization results for the clockwise and counterclockwise moving adversary are displayed in Table \ref{table:3}. 

\begin{table}[!ht]
\centering
\begin{tabular}{ |p{3.3cm}||p{3.3cm}|p{3.3cm}|  }
 \hline
 \multicolumn{3}{|c|}{$5\times 5$ Environment - Online Optimization Results} \\
 \hline
Adversary Movement  & Clockwise & Counterclockwise \\
 \hline
\# of Successes   &  50   & 50\\
Average Reward & 294	& 294\\
Average Regret & 0	&	0\\
Time(secs) & 48.89	&	47.32\\
 \hline
\end{tabular}
 \caption{$5\times 5$ Environment: Online Optimization Results}
 \label{table:3}
\end{table}

From the results in Table \ref{table:3}, we can see that the online optimization framework works perfectly and efficiently when the adversary moves deterministically. When the adversary moves randomly, we are unable to succeed one hundred percent of the time with as little as 8 time step observations. We need to increase the number of observations before allowing our agent to begin the game in order to ensure success. The results in Table \ref{table:4} show the online optimization results after collecting 10, 25, 50, and 75 observations on a randomly moving adversary. 

\begin{table}[!ht]
\centering
\begin{tabular}{ |p{3.2cm}||p{1.67cm}|p{1.67cm}|p{1.67cm}|p{1.67cm}|  }
 \hline
 \multicolumn{5}{|c|}{$5\times 5$ Environment - Online Optimization Results} \\
 \hline
Adversary Movement & Random& Random& Random& Random \\
 \hline
 \# of Observations & 10 & 25 & 50 & 75\\
\# of Successes   &  18 & 27 & 36 & 50\\
Average Reward & 294 & 293.04 & 290.61 & 278.88\\
Average Regret & 0	& .96 &	3.39 & 15.12\\
Time(secs) & 29.27	&	39.71	&	69.74 & 69.69\\
 \hline
\end{tabular}
 \caption{$5\times 5$ Environment: Online Optimization Results}
 \label{table:4}
\end{table}

\begin{figure}[!htp]
    \centering
    \includegraphics[width=\textwidth]{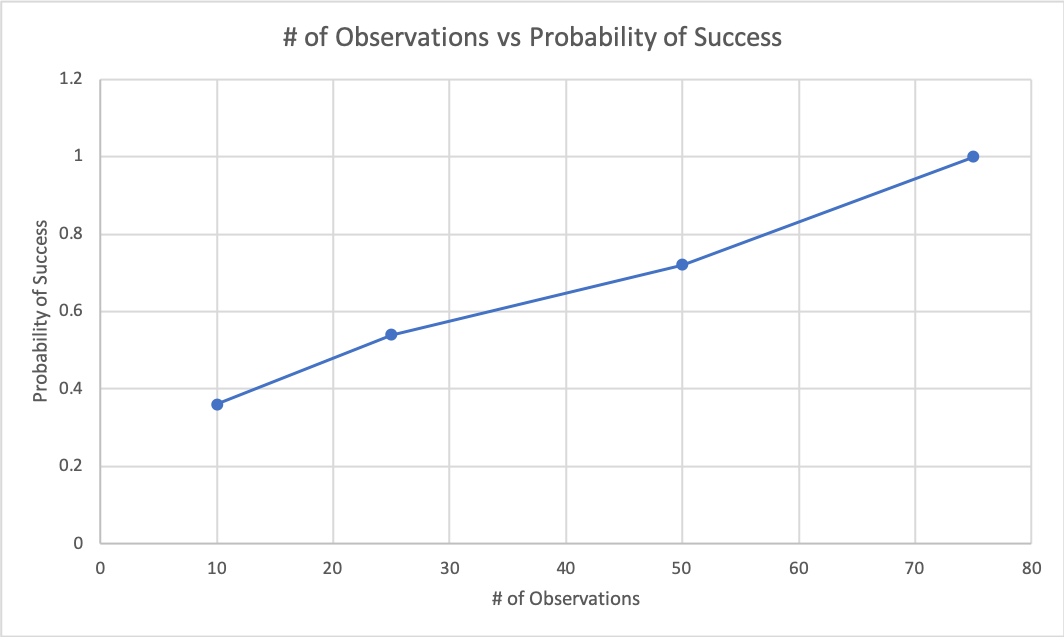}
    \caption{$5\times 5$ Online Optimization Randomly Moving Adversary Results}
    \label{fig:5x5linechart}
\end{figure}

We can see from Table \ref{table:4} that we do not successfully complete all 50 runs with online optimization until we are able to observe the adversary for 75 time steps. Another thing to note is that while success rate is low with fewer observations, there appears to be higher reward. This is due to the fact that when the agent does not have accurate knowledge of how the adversary will move, it speeds through the grid using the shortest path and often gets captured. When the agent is able to avoid the adversary by chance, it is able to complete the game using the shortest path and achieve a high reward. As the agent has more accurate knowledge of how the adversary will move, it will take more steps in order to avoid the adversary and therefore take a longer path to complete the game, resulting in a lower reward. A line chart plotting the number of observations against the probability of success is shown in Figure \ref{fig:5x5linechart}.

\begin{figure}[!htp]
    \centering
    \includegraphics[width=4cm]{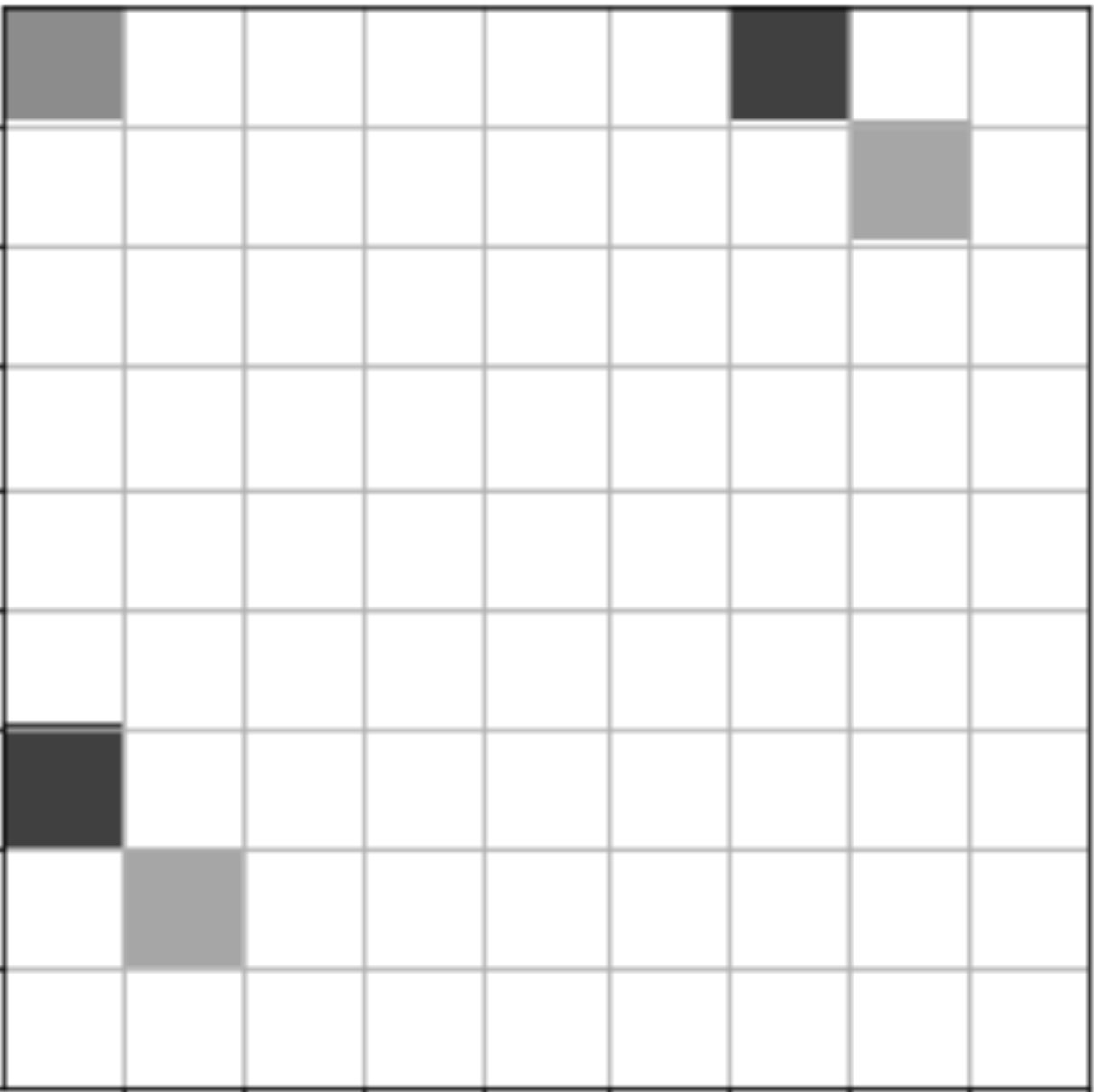}
    \caption{$9\times 9$ Example Initial Setup}
    \label{fig:9x9setup}
\end{figure}

Now, let us expand our experiment by increasing the grid size to $9\times 9$ and allowing there to be 2 rewards and 2 adversaries. We will place the adversaries far from each other and start the adversaries again in the upper left cell adjacent to each reward. With a bigger environment and two rewards, the optimal reward now is 475. Figure \ref{fig:9x9setup} shows the set up of the game. 

{\color{black} We revisit our $\varepsilon$-greedy tabular $Q$-Learning model and examine how it performs against the new $9\times 9$ grid environment with two adversaries.  Given the larger state-space, we increased the number of epochs dedicated to training our UAV from 1,000 to 2,500.  The additional epochs were necessary to ensure that the UAV would visit a significant portion of the available environment in order to avoid generating a sparse $Q$-Learning look-up table.  As we did on the $5\time 5$ grid environment, we ran the 2,500 epoch simulation ten times for each of the three adversary movement patterns: clockwise, counter-clockwise, and random.

As compared to the results of the $5\times 5$ environment, the $\varepsilon$-greedy tabular method struggled to find an effective feasible policy for the complete $9\times 9$ grid environment.  In all of our scenarios the UAV was unable to learn a feasible policy that could locate both rewards and exit the environment while also avoiding both adversaries.  At best, the agent was able to learn an effective policy for capturing only one of the rewards, either the top-right or bottom-left reward.  Unfortunately, once the initial award was captured the agent was unable to learn a policy for traversing to the opposite side of the grid and capturing the second award.  These results are not entirely surprising and are consistent with the theory that the $\varepsilon$-greedy tabular $Q$-learning will breakdown quickly when applied to larger state-space environments due to the curse of dimensionality that was discussed earlier.  

We were able to gain moderate improvements in our $Q$-learning look-up table results by doubling the number of dedicated training epochs from 2,500 to 5,000. However, even with the increase in the number of training epochs, the UAV was still unable to generate a feasible policy for successfully completing the entire grid environment exercise.  Furthermore, although there was noticeable improvement to the UAVs ability to find the initial award, this marginal improvement was gained at a consider cost to the time required to run all 5,000 epochs.  A summary of the $\varepsilon$-greedy tabular $Q$-Learning model results for the $9\times 9$ grid environment are shown in Table \ref{table:5}.

\renewcommand*{\arraystretch}{1.3}
\begin{table}[!ht]
\centering
\begin{tabular}{ |p{5.5cm}||p{1.5cm}|p{1.5cm}|p{1.5cm}|  }
 \hline
 \multicolumn{4}{|c|}{$9\times 9$ Environment - $\epsilon$-Greedy Tabular Method} \\
 \hline
Adversary Movements  & Clockwise & Counter- & Random\\
 & & clockwise & \\
 \hline Total number of training epochs: 2,500 & & & \\
Number of successful single award policies & 1 & 2 & 0\\ 
One $Q$-value away from a successful single award policy & 5 & 6 & 5\\
Avg. time(secs.) & 45.86 & 47.32 & 44.31\\  
\hline
Total number of training epochs: 5,000 & 	& 	&	\\
Number of successful single award policies & 7 & 6 & 3  \\
One $Q$-Value away from a successful single award policy &  2 & 2 & 1  \\
Avg. time(secs.) & 7,537.23 & 7,333.12 & 6,673.18 \\
\hline
 \end{tabular}
 \caption{$9\times 9$ Environment: $\epsilon$-Greedy Tabular $Q$-Learning Results}
 \label{table:5}
\end{table}

\renewcommand*{\arraystretch}{1}
}

We now examine the results of our $9\times 9$ grid environment using Deep $Q$-Learning.  Again, we received much different results when the environment size is increased. 

\begin{table}[!ht]
\centering
\begin{tabular}{ |p{3cm}||p{2.3cm}|p{2.3cm}|p{2.3cm}|  }
 \hline
 \multicolumn{4}{|c|}{$9\times 9$ Environment - Deep $Q$-Learning Results} \\
 \hline
Adversary Movement  & Clockwise & Counterclockwise & Random\\
 \hline
\# of Successes   &  7   & 4 & 0 \\
Average Reward & 322.43	& 342.5	&	0\\
Average Regret & 152.57	&132.5&	475\\
Time (days) & 8.75	&	9	&	1.69\\
 \hline
\end{tabular}
 \caption{$9\times 9$ Environment: Deep-$Q$ Learning Results}
 \label{table:6}
\end{table}

From Table \ref{table:6}, we can see that the algorithm took much longer time to run and achieve minimal success. This is due to the fact that our agent has no idea where the rewards are located. Once our agent has reached one reward and received a positive response, it has little incentive to go experiment crossing the entire grid in order to reach another reward that may not be there. The only reason that the Deep $Q$-Learning algorithm has any success is due to the exploration nature of the algorithm. Even with a 1000 epochs, or 1000 tries, for each of the 50 times we ran our Deep $Q$-Learning algorithm, our agent was still unable to successfully complete the game a single time when the adversary moved randomly. 

When we tried completing the $9\times 9$ game environment using online optimization, we saw similar results to the $5\times 5$ game environment, but with much longer computational time. For the clockwise and counterclockwise moving adversaries, 8 time step observations was once again enough to ensure success as we can see in Table \ref{table:7}. 

\begin{table}[!ht]
\centering
\begin{tabular}{ |p{3.3cm}||p{3.3cm}|p{3.3cm}|  }
 \hline
 \multicolumn{3}{|c|}{$9\times 9$ Environment - Online Optimization Results} \\
 \hline
Adversary Movement  & Clockwise & Counterclockwise \\
 \hline
\# of Successes   &  50   & 50\\
Average Reward & 475	& 475\\
Average Regret & 0	&	0\\
Time(secs) & 1472.28	&	2250.28\\
 \hline
\end{tabular}
 \caption{$9\times 9$ Environment: Online Optimization Results}
 \label{table:7}
\end{table}

We can see that with 8 time step observations, online optimization is able to achieve the optimal reward and always succeed when the adversary moves deterministically. When the adversary moves randomly, we see similar results to the $5\times 5$ case, in that success is not ensured until about after collecting observations for 75 time steps. 

\begin{table}[!ht]
\centering
\begin{tabular}{ |p{3.2cm}||p{1.67cm}|p{1.67cm}|p{1.67cm}|p{1.67cm}|  }
 \hline
 \multicolumn{5}{|c|}{$9\times 9$ Environment - Online Optimization Results} \\
 \hline
Adversary Movement & Random & Random & Random & Random \\
 \hline
 \# of Observations & 10 & 25 & 50 & 75\\
\# of Successes   &  11 & 17 & 24 & 50\\
Average Reward & 475 & 472.88 & 465.5 & 453.4\\
Average Regret & 0	& 2.12 & 9.5 & 21.6\\
Time(hours) & 0.26	& 0.33	& 5.79 & 8.75\\
 \hline
\end{tabular}
 \caption{$9\times 9$ Environment: Online Optimization Results}
 \label{table:8}
\end{table}

\begin{figure}[!ht]
    \centering
    \includegraphics[width=\textwidth]{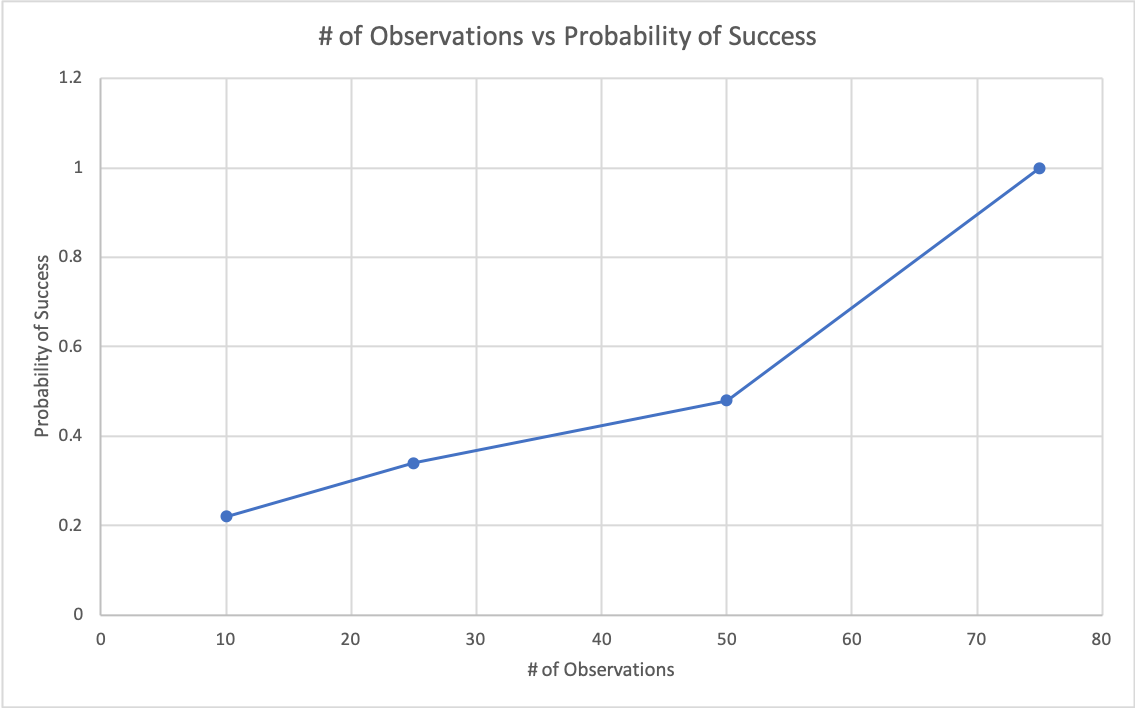}
    \caption{$9\times 9$ Online Optimization Randomly Moving Adversary Results}
    \label{fig:6}
\end{figure}

From Table \ref{table:8}, we see very similar trends between the $9\times 9$ and the $5\times 5$ case in that with fewer time step observations, while the probability of success increases with more time step observations, the average reward of successful runs go down. A line chart plotting the number of observations against the probability of success is shown in Figure \ref{fig:6}.

\section{Conclusion}

In this paper, we were able to learn a great deal on how the three algorithms perform when faced with our specific problem setup. In the $5\times 5$ environment with only a single adversary, we saw that the $\epsilon$-Greedy Tabular method worked reasonable well for the clockwise and counterclockwise adversary.  Also, although the algorithm did not perform as well when faced with a stochastic moving adversary, a closer look at the results showed that if the algorithm could adjust its learning for one critical $Q$ value its success rate would be on par with the counterclockwise scenario.  Regardless, for the $5\times 5$ environment, the tabular method is extremely fast, only taking roughly a second to complete 500 epochs and learn 200 $Q$-values.  However, for the much larger $9\times 9$ environment, we look to alternative methods such as Deep-$Q$ learning and online optimization.     

In the $5\times 5$ Grid environment, the Deep-$Q$ Learning and online optimization algorithms performed relatively well with quick computational times and low regret. However, once we increased the game environment and placed two sets of rewards and adversaries far from each other, Deep $Q$-Learning performed much worse than online optimization. This is due to the exploratory nature of our Deep $Q$-Learning algorithm as it has zero knowledge of the environment and learn everything through trial and error. Our online optimization algorithm may not know initially how the adversary moves, but it is given the chance to observe the adversary for a certain number of time steps and has complete knowledge of the reward and adversary locations. While it is unfair to compare the two algorithms based on our results because they begin with different amounts of knowledge of their operating environment, we were still able to gain a lot of insight into the two algorithms in regards to a game grid environment. 

When we increase the size of the game environment and the number of rewards, it appears that the Deep $Q$-Learning algorithm has trouble locating all of the rewards and completing the game as there is a penalty for each additional step it must take. Once it has found one reward, it has little motivation to keep exploring further and further in hopes of finding an additional reward. Online optimization works well in our experiments, but the computational time increases significantly when we increase our game environment as the algorithm has much more computations. It is interesting to note that in both our smaller and bigger experiment grids, it took 75 time steps before our algorithm was able to successfully predict the movement of the adversaries and complete the game every time when the adversary moves randomly.

The experiments we ran were initial experiments, and there are much more additional experiments worth exploring. Some future work may involve adjusting the parameters of our Deep $Q$-Learning algorithm and online optimization algorithms to increase efficiency and effectiveness, incorporating parallel processing into our codes to decrease computational time, and doing more experiments with different game environments, rewards, and adversaries to learn more about the two algorithms. 

{\color{black}In general, finding a Nash equilibrium is PPAD-complete for two players' game \citep{nashequi}.  Therefore, it is intractable even for linear online optimization problem.  \cite{10.1007/978-3-642-02930-1_35} showed that for the zero-sum game on a network, a Nash equilibrium can be found efficiently with linear-programming. However, in our game, the agent's actions and the adversary's actions are not symmetric.  Therefore, it is an open question whether there exists a Nash equilibrium and, if there exists a Nash equilibrium, whether it is efficient to find one or not.  }

\section{Experimental Reproducibility}
\label{reprod}

{\color{black} All codes used for computations and experimentation are posted online at \url{https://github.com/benliu31492/Solving-reward-collecting-problems-with-UAVS-a-comparison-of-online-optimization-and-Q-Learning} with all necessary documentation and prerequisites to run. 
This GitHub repository contains the codes for all three approaches presented in our manuscript ``Solving reward-collecting problems with UAVs: a comparison of online optimization and $Q$-learning''. Specifically, there are three sets of codes for the three methods described in the paper: \begin{enumerate}
    \item Deep $Q$-Learning,
    \item Online Optimization, and 
    \item $\epsilon$-greedy tabular $Q$-Learning.
\end{enumerate} For each method, we post codes that explore either a $5 \times 5$ grid environment with one adversary or a $9 \times 9$ grid environment two adversaries.
To successfully compile and run, the interested reader would need Tensorflow 2.5.0, Keras 2.4.3, OpenAI Gym 0.18.3, and the Gurobi {\tt Python} packages.
Please refer to this Github page to perform any further computations.
}






\begin{acknowledgements}
{\color{black}The authors would like to thank CAPT (Ret) Jeff Kline at Naval Postgraduate School and Dr.~Timothy Bentley at the Office of Naval Research for sharing information on the problem of Autonomous Casualty Evacuation.}
\end{acknowledgements}
\section*{Declarations}

{\bf Ethical Approval}:
This paper does not contain any studies with human participants or animals performed by any of the authors. 

\noindent
{\bf Consent to Participate}: 
Not applicable.

\noindent
{\bf Consent to Publish}:
Not applicable.

\noindent
{\bf Authors Contributions}:
\begin{enumerate}
    \item Yixuan Liu: Implemented the Deep $Q$-Learning and online optimization framework and conducted all computational experiments. 
    \item Chrysafis Vogiatzis: Modeled the online optimization framework and supervised Y.L. for implementation and computational experiments.
    \item Ruriko Yoshida: Designed the game and coordinated the research.   
    \item Erich Morman: Modeled and implemented the $\varepsilon$-greedy tabular $Q$-Learning.  Additionally conducted computational experiments using $\epsilon$-Q learning. 
\end{enumerate}
All authors discussed the results and contributed to the final manuscript.
 
\noindent
{\bf Funding}:
R.Y. is partially supported by NSF DMS 1916037 and Consortium for Robotics and Unmanned Systems Education and Research (CRUSER).

\noindent
{\bf Competing Interests}: 
The authors declare that they have no conflict of interest.

\noindent
{\bf Availability of data and materials}: 
All data and materials are available upon request. \textcolor{black}{The codes are also available on GitHub (see Section \ref{reprod}).}

\bibliographystyle{spbasic_updated}      
\bibliography{sample}   

\begin{thebibliography}{32}
\providecommand{\natexlab}[1]{#1}
\providecommand{\url}[1]{{#1}}
\providecommand{\urlprefix}{URL }
\expandafter\ifx\csname urlstyle\endcsname\relax
  \providecommand{\doi}[1]{DOI~\discretionary{}{}{}#1}\else
  \providecommand{\doi}{DOI~\discretionary{}{}{}\begingroup
  \urlstyle{rm}\Url}\fi
\providecommand{\eprint}[2][]{\url{#2}}

\bibitem[{Abadi et~al.(2015)Abadi, Agarwal, Barham, Brevdo, Chen, Citro,
  Corrado, Davis, Dean, Devin, Ghemawat, Goodfellow, Harp, Irving, Isard, Jia,
  Jozefowicz, Kaiser, Kudlur, Levenberg, Man\'{e}, Monga, Moore, Murray, Olah,
  Schuster, Shlens, Steiner, Sutskever, Talwar, Tucker, Vanhoucke, Vasudevan,
  Vi\'{e}gas, Vinyals, Warden, Wattenberg, Wicke, Yu, and Zheng}]{tensorflow}
Abadi M, Agarwal A, Barham P, Brevdo E, Chen Z, Citro C, Corrado GS, Davis A,
  Dean J, Devin M, Ghemawat S, Goodfellow I, Harp A, Irving G, Isard M, Jia Y,
  Jozefowicz R, Kaiser L, Kudlur M, Levenberg J, Man\'{e} D, Monga R, Moore S,
  Murray D, Olah C, Schuster M, Shlens J, Steiner B, Sutskever I, Talwar K,
  Tucker P, Vanhoucke V, Vasudevan V, Vi\'{e}gas F, Vinyals O, Warden P,
  Wattenberg M, Wicke M, Yu Y, Zheng X (2015) {TensorFlow}: Large-scale machine
  learning on heterogeneous systems

\bibitem[{Belmega et~al.(2018)Belmega, Mertikopoulos, Negrel, and
  Sanguinetti}]{online_formulation}
Belmega V, Mertikopoulos P, Negrel R, Sanguinetti L (2018) Online convex
  optimization and no-regret learning: Algorithms, guarantees and applications.
  arXiv preprint arXiv:1804.04529

\bibitem[{Board et~al.(2005)Board, Council et~al.}]{NAP11379}
Board NS, Council NR, et~al. (2005) Autonomous vehicles in support of naval
  operations. National Academies Press, Washington, DC

\bibitem[{Bubeck(2011)}]{bubeck_2011}
Bubeck S (2011) Introduction to online optimization. Lecture, Introduction to
  Veterinary Studies, May 2, Department of Dragon Husbandry, Charlatan State
  University, Monogahela, WV

\bibitem[{Burkov(2019)}]{machinelearning}
Burkov A (2019) The hundred-page machine learning book, vol~1. Andriy Burkov,
  Quebec City, Canada

\bibitem[{Carta et~al.(2020)Carta, Ferreira, Podda, Recupero, and
  Sanna}]{carta2020multi}
Carta S, Ferreira A, Podda AS, Recupero DR, Sanna A (2020) Multi-dqn: An
  ensemble of deep q-learning agents for stock market forecasting. Expert
  Systems with Applications 164:113,820

\bibitem[{Chen and Deng(2006)}]{nashequi}
Chen X, Deng X (2006) Settling the complexity of two-player nash equilibrium.
  In: 2006 47th Annual IEEE Symposium on Foundations of Computer Science
  (FOCS'06), pp 261--272, \doi{10.1109/FOCS.2006.69}

\bibitem[{Chollet et~al.(2015)}]{chollet2015keras}
Chollet F, et~al. (2015) Keras. \url{https://keras.io}

\bibitem[{Darken et~al.(1992)Darken, Chang, Moody et~al.}]{Darken}
Darken C, Chang J, Moody J, et~al. (1992) Learning rate schedules for faster
  stochastic gradient search. In: Neural networks for signal processing,
  Citeseer, vol~2

\bibitem[{Daskalakis and Papadimitriou(2009)}]{10.1007/978-3-642-02930-1_35}
Daskalakis C, Papadimitriou CH (2009) On a network generalization of the minmax
  theorem. In: Albers S, Marchetti-Spaccamela A, Matias Y, Nikoletseas S,
  Thomas W (eds) Automata, Languages and Programming, Springer Berlin
  Heidelberg, Berlin, Heidelberg, pp 423--434

\bibitem[{{Defense Systems Information Analysis Center}(2020)}]{ACE2}
{Defense Systems Information Analysis Center} (2020) {Autonomous Unmanned
  Vehicles for Casualty Evacuation Support}.
  \url{https://www.dsiac.org/services/technical-inquiries/notable-ti/autonomous-unmanned-vehicles-for-casualty-evacuation-support/}

\bibitem[{Faust et~al.(2017)Faust, Palunko, Cruz, Fierro, and Tapia}]{faust}
Faust A, Palunko I, Cruz P, Fierro R, Tapia L (2017) Automated aerial suspended
  cargo delivery through reinforcement learning. Artificial Intelligence
  247:381--398

\bibitem[{Gosavi(2003)}]{Gosavi_2}
Gosavi A (2003) Simulation-Based Optimization: Parametric Optimization
  Techniques and Reinforcement Learning. Kluwer Academic Publishers, Boston, MA

\bibitem[{Gosavi(2009)}]{Gosavi}
Gosavi A (2009) Reinforcement learning: A tutorial survey and recent advances.
  INFORMS Journal on Computing 21(2):178--192

\bibitem[{{Gurobi Optimization}(2020)}]{gurobi}
{Gurobi Optimization} (2020) Gurobi optimizer reference manual

\bibitem[{He et~al.(2013)He, Goeckel, Raghavendra, and Towsley}]{he2013endhost}
He T, Goeckel D, Raghavendra R, Towsley D (2013) Endhost-based shortest path
  routing in dynamic networks: An online learning approach. In: 2013
  Proceedings IEEE INFOCOM, IEEE, pp 2202--2210

\bibitem[{Hoehn and Sayler(2020)}]{hoehn_sayler_2020}
Hoehn JR, Sayler KM (2020) Department of defense counter-unmanned aircraft
  systems

\bibitem[{Ingrand and Ghallab(2017)}]{survey}
Ingrand F, Ghallab M (2017) Deliberation for autonomous robots: A survey.
  Artificial Intelligence 247:10--44

\bibitem[{Li and Hoi(2014)}]{li2014online}
Li B, Hoi SC (2014) Online portfolio selection: A survey. ACM Computing Surveys
  (CSUR) 46(3):1--36

\bibitem[{Maas et~al.(2013)Maas, Hannun, and Ng}]{leakyrelu}
Maas AL, Hannun AY, Ng AY (2013) Rectifier nonlinearities improve neural
  network acoustic models. In: in ICML Workshop on Deep Learning for Audio,
  Speech and Language Processing

\bibitem[{Mnih et~al.(2013)Mnih, Kavukcuoglu, Silver, Graves, Antonoglou,
  Wierstra, and Riedmiller}]{Mnih}
Mnih V, Kavukcuoglu K, Silver D, Graves A, Antonoglou I, Wierstra D, Riedmiller
  A (2013) Playing atari with deep reinforcement learning. CoRR abs/1312.5602

\bibitem[{Powell(2011)}]{Powell}
Powell WB (2011) Approximate Dynamic Programming: Solving the Curses of
  Dimensionality. John Wiley \& Sons, Inc., Hoboken, NJ

\bibitem[{Qiao et~al.(2018)Qiao, Wang, Li, and Chen}]{qiao2018adaptive}
Qiao J, Wang G, Li W, Chen M (2018) An adaptive deep q-learning strategy for
  handwritten digit recognition. Neural Networks 107:61--71

\bibitem[{Reddi et~al.(2018)Reddi, Kale, and Kumar}]{adam}
Reddi S, Kale S, Kumar S (2018) On the convergence of adam and beyond. In:
  International Conference on Learning Representations

\bibitem[{Shu(2014)}]{Shu}
Shu C (2014) Google acquires artificial intelligence startup deepmind for more
  than \$500m

\bibitem[{Sutton and Barto(2018)}]{Sutton}
Sutton R, Barto A (2018) Reinforcement learning an introduction. The MIT Press,
  Cambridge, MA

\bibitem[{{The Robot Report}(2020)}]{ACE1}
{The Robot Report} (2020) {Autonomous Casualty Extraction program awarded to
  RE2 Robotics by U.S. Army}.
  \url{https://www.therobotreport.com/autonomous-casualty-extraction-funding-awarded-re2-robotics-army/}

\bibitem[{Wang et~al.(2006)Wang, Guan, and Wang}]{wang2006svm}
Wang Q, Guan Y, Wang X (2006) Svm-based spam filter with active and online
  learning. In: TREC, Citeseer

\bibitem[{Watkins and Dayan(1992)}]{Watkins}
Watkins CJ, Dayan P (1992) Technical note. Reinforcement Learning pp 55--68

\bibitem[{Williams et~al.(2019)Williams, Sebastian, and
  Ben-Tzvi}]{williams2019review}
Williams A, Sebastian B, Ben-Tzvi P (2019) Review and analysis of search,
  extraction, evacuation, and medical field treatment robots. Journal of
  Intelligent \& Robotic Systems 96(3):401--418

\bibitem[{Zafrany(2017)}]{zafrany}
Zafrany S (2017) Deep reinforcement learning the tour de flags test case.
  \url{https://www.samyzaf.com/ML/tdf/tdf.html}

\bibitem[{Zhang et~al.(2017)Zhang, Lin, Yang, Chen, and Li}]{zhang2017energy}
Zhang Q, Lin M, Yang LT, Chen Z, Li P (2017) Energy-efficient scheduling for
  real-time systems based on deep q-learning model. IEEE transactions on
  sustainable computing 4(1):132--141

\end{thebibliography}

\end{document}